\pdfoutput=1

\documentclass[11pt]{article}

\usepackage[preprint]{acl}

\usepackage{times}
\usepackage{latexsym}
\usepackage{amsfonts}
\usepackage{amsmath}

\usepackage[T1]{fontenc}

\usepackage[utf8]{inputenc}

\usepackage{microtype}

\usepackage{inconsolata}
\usepackage{url}

\usepackage{booktabs}
\usepackage{amssymb}
\usepackage{graphicx}
\usepackage{xspace}
\usepackage{multirow}
\usepackage{multicol}
\usepackage{subcaption}
\usepackage{xcolor}
\usepackage{tcolorbox}
\tcbuselibrary{breakable} 
\newtcolorbox[auto counter, number within=section, list type=subsubsection, list inside=toc]{sectionbox}[1]{
colback=white, colframe=black, 
colbacktitle=white!80!gray, coltitle=black, 
fonttitle=\bfseries, title={Comment \thetcbcounter}, list entry={Comment \thetcbcounter\quad}, 
breakable, 
before upper={\parindent10pt\noindent},  
left = 1mm, 
    right = 1mm,
    top = 1mm,
    bottom = 1mm,
}
\newcommand{\name}{LLaMA-Omni 2\xspace}

\title{\name: LLM-based Real-time Spoken Chatbot with Autoregressive Streaming Speech Synthesis}

\author{
    Qingkai Fang$^{1,3}$,
    Yan Zhou$^{1,3}$,
    Shoutao Guo$^{1,3}$,
    Shaolei Zhang$^{1,3}$,
    Yang Feng$^{1,2,3}$\thanks{Corresponding author: Yang Feng.} \\
    \textsuperscript{\rm1}Key Laboratory of Intelligent Information Processing \\ Institute of Computing Technology, Chinese Academy of Sciences (ICT/CAS) \\
    \textsuperscript{\rm2}Key Laboratory of AI Safety, Chinese Academy of Sciences \\
    \textsuperscript{\rm3}University of Chinese Academy of Sciences, Beijing, China \\
    {$\;\:$\texttt{\{\href{mailto:fangqingkai21b@ict.ac.cn}{fangqingkai21b},\href{mailto:fengyang@ict.ac.cn}{fengyang}\}@ict.ac.cn}}
}

\begin{document}
\maketitle
\begin{abstract}
Real-time, intelligent, and natural speech interaction is an essential part of the next-generation human-computer interaction. Recent advancements have showcased the potential of building intelligent spoken chatbots based on large language models (LLMs). In this paper, we introduce \name, a series of speech language models (SpeechLMs) ranging from 0.5B to 14B parameters, capable of achieving high-quality real-time speech interaction. \name is built upon the Qwen2.5 series models, integrating a speech encoder and an autoregressive streaming speech decoder. Despite being trained on only 200K multi-turn speech dialogue samples, \name demonstrates strong performance on several spoken question answering and speech instruction following benchmarks, surpassing previous state-of-the-art SpeechLMs like GLM-4-Voice, which was trained on millions of hours of speech data.\footnote{Code: \url{https://github.com/ictnlp/LLaMA-Omni2}\\ \hspace*{1.5em} Audio Samples: \url{https://llama-omni2.github.io/}}
\end{abstract}

\section{Introduction}
Speech, as a critical interface for human-computer interaction, can significantly enhance both interaction efficiency and user experience~\citep{10.1093/iwc/iwz016}. In recent years, as large language models (LLMs) like ChatGPT~\citep{chatgpt} have demonstrated outstanding performance across various fields, speech interactions with LLMs have attracted widespread attention from both academia and industry. For instance, GPT-4o~\citep{gpt4o} enables \textit{real-time}, \textit{intelligent}, and \textit{natural} speech interaction between users and LLMs, heralding the advent of a new generation of human-computer interaction paradigms.

To develop a spoken chatbot similar to GPT-4o, the traditional approach typically employs a cascaded pipeline comprising an automatic speech recognition (ASR) model, an LLM, and a text-to-speech (TTS) model. While this method is relatively straightforward to implement, it suffers from several notable limitations. First, errors can accumulate across the different stages of the pipeline. Second, the overall response latency tends to be high due to the sequential processing of multiple models. Third, the system struggles to capture paralinguistic information present in the input speech. To address these limitations, end-to-end speech language models (SpeechLMs) have gradually gained more attention, using a single unified model to handle the entire process from speech input to output. Overall, end-to-end SpeechLMs can be categorized into two types: \textit{native} and \textit{modular}. Native SpeechLMs typically discretize speech into tokens and employ a GPT-style decoder-only Transformer~\citep{gpt1} to model both speech and text within a unified language model~\citep{zhang-etal-2023-speechgpt, rubenstein2023audiopalm, twist}. A key advantage of this architecture is its ability to leverage vast amounts of unsupervised speech data for pretraining, making it easier to scale up in terms of model parameters and data size. This can potentially result in emergent capabilities, such as more human-like speech expressiveness~\citep{glm4voice, speechgpt2preview}. However, native SpeechLMs typically require large-scale speech datasets (e.g., millions of hours) for pretraining~\citep{zeng2024scaling, kyutai2024moshi}, which presents challenges in data collection and training costs, and may also lead to catastrophic forgetting of the model's text capabilities. In contrast, modular SpeechLMs incorporate a speech encoder and a speech decoder around the LLM to handle speech understanding and generation~\citep{fang2025llamaomni, xiong2024freeze}. The advantage of this approach is its ability to leverage the inherent capabilities of each module, requiring only small-scale fine-tuning (e.g., a few hundred or thousand hours of speech data) to align the modules. This enables the model to acquire speech interaction capabilities at a relatively low cost, while retaining most of its original capability. Moreover, modular SpeechLMs can typically generate speech guided by textual output, ensuring the intelligence of the generated speech.

In addition to the intelligence of speech, real-time responsiveness and naturalness are also crucial characteristics of spoken chatbots. LLaMA-Omni~\citep{fang2025llamaomni} uses a non-autoregressive (NAR) streaming speech decoder to enable synchronized generation of speech and text, ensuring extremely low response latency. However, due to the limitations of non-autoregressive models in modeling capacity, the generated speech is often less natural and fluent. Freeze-Omni~\citep{xiong2024freeze} combines both NAR and autoregressive (AR) models for speech generation, resulting in higher naturalness of the generated speech. However, it can only achieve sentence-level streaming speech generation through a simple sentence-split strategy, which prevents it from achieving very low response latency. To address these challenges, in this paper, we introduce \name, a series of modular SpeechLMs ranging from 0.5B to 14B. \name adopts Qwen2.5-0.5B/1.5B/3B/7B/14B-Instruct models~\citep{qwen2.5} as the base LLM, and uses Whisper's encoder~\citep{whisper} as the speech encoder. For the speech decoder, inspired by the state-of-the-art streaming speech synthesis model CosyVoice 2~\citep{du2024cosyvoice}, it first includes an autoregressive text-to-speech language model initialized with Qwen2.5-0.5B, which generates speech tokens from the LLM output and achieves streaming generation through alternating read and write operations. The speech tokens are then passed through a chunk-aware causal flow matching model~\citep{lipman2023flow} to generate the mel spectrogram in a streaming manner. To train the model, we synthesize 200K multi-turn speech-to-speech dialogue samples with diverse input voices and a uniform output voice. Experimental results show that \name achieves outstanding performance on spoken question answering and speech instruction following tasks in both speech-to-text and speech-to-speech settings, outperforming both LLaMA-Omni and the native SpeechLM GLM-4-Voice~\citep{glm4voice}, which was trained on millions of hours of speech data. We also conducted detailed ablation studies on factors such as LLM parameter size, training data scale, speech decoder pretraining, and read-write strategy, to better understand the impact of these factors on the overall system performance.

\section{Model: \name}
\begin{figure*}[t]
    \centering
    \includegraphics[width=\textwidth]{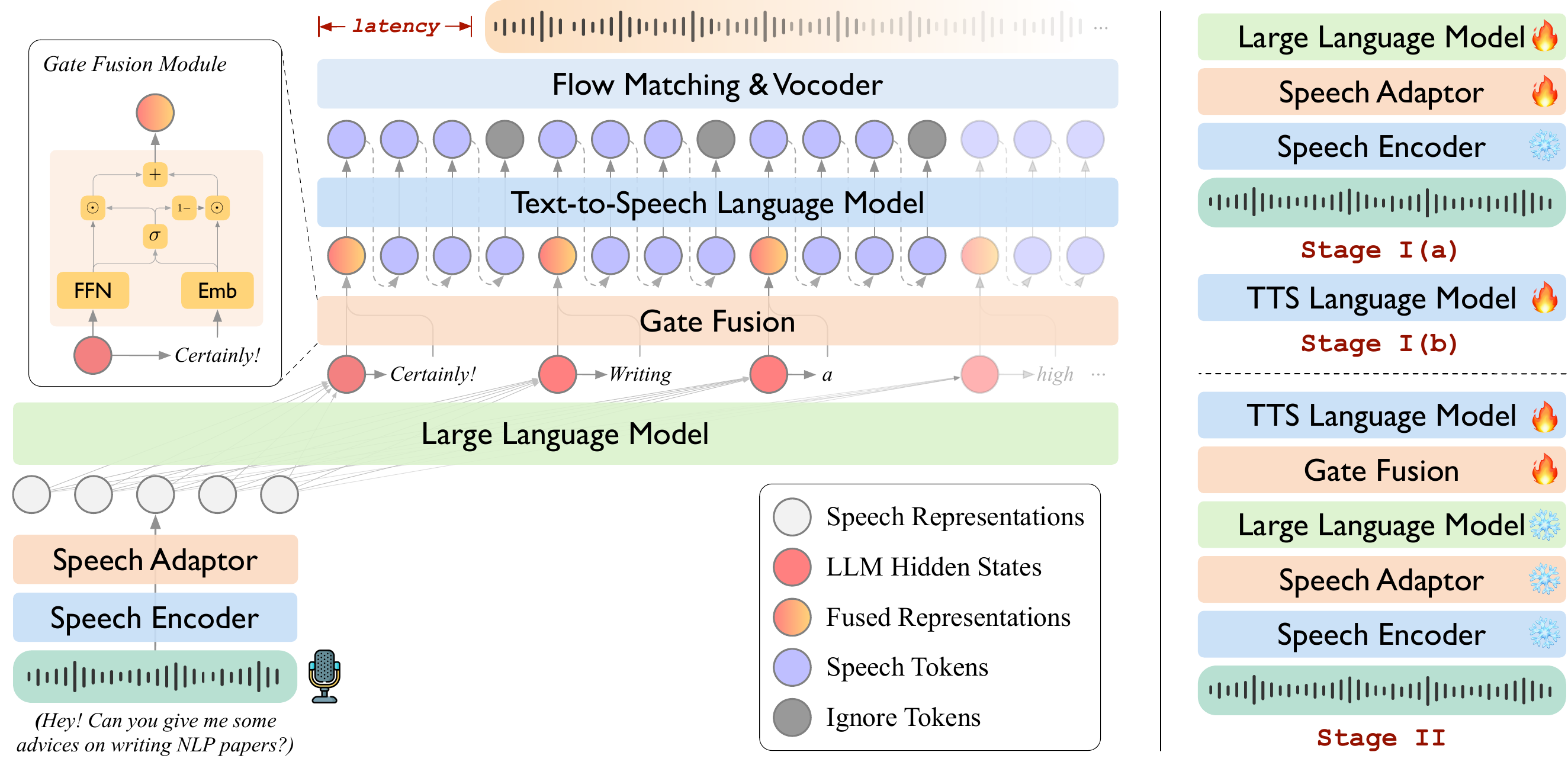}
    \caption{\textbf{Left:} Model architecture of \name. \textbf{Right:} Illustration of the two-stage training strategy.}
    \label{fig:model}
\end{figure*}

In this section, we introduce the model architecture of \name. As shown in Figure~\ref{fig:model}, the core of \name is an LLM, for which we use the Qwen2.5 series models~\citep{qwen2.5} due to their strong performance across various benchmarks. Next, we will describe how we equip the LLM with speech understanding and streaming speech generation capabilities. In the following, we use $\mathcal{M}_\text{LLM}$ to denote the LLM. For a single-turn instruction-response pair, we denote the speech instruction as $X$, and the text and speech responses as $Y^T$ and $Y^S$, respectively. 

\subsection{Speech Understanding}
To enable speech understanding, we incorporate a speech encoder and a speech adapter before the LLM, similar to LLaMA-Omni~\citep{fang2025llamaomni}. Specifically, we use the encoder of Whisper-large-v3~\citep{whisper} as the speech encoder, which converts the input speech into a sequence of representations. The encoded representations are then passed into the speech adapter, which consists of a downsampling module and a feed-forward network (FFN). The downsampling module concatenates every $k$ consecutive frames along the feature dimension, and the concatenated representations are further encoded by the FFN. The final output representation is then input into the LLM.

\subsection{Streaming Speech Generation}
To equip the model with streaming speech generation capabilities, we adopt a paradigm similar to CosyVoice 2~\citep{du2024cosyvoice}. First, the speech response is converted into discrete tokens using a supervised semantic speech tokenizer. Then, an autoregressive text-to-speech language model is employed to model the streaming generation from the LLM output to speech tokens. Finally, a causal flow matching model converts speech tokens into the mel spectrogram in a streaming manner.

\paragraph{Speech Tokenizer}
The speech tokenizer is implemented by inserting a finite scalar quantization (FSQ) module~\citep{mentzer2024finite} into the encoder of SenseVoice-Large ASR model~\citep{an2024funaudiollm}. This module first projects the intermediate representations to a low-rank space and discretizes them through a rounding operation. Ultimately, the speech response $Y^S$ is converted into a token sequence $Y^U = [y^U_1, \dots, y^U_M]$, with 25 tokens per second, where each token $y^U_i \in \{ K \in \mathbb{N} \mid 0 \leq K < 6561 \}$. We use the pretraiend speech tokenizer in CosyVoice 2. 

\paragraph{Text-to-Speech Language Model}
After converting the speech response into discrete tokens, we use a decoder-only Transformer~\citep{vaswani2017attention} to model the conditional language model from the LLM output to the speech tokens, denoted as $\mathcal{M}_\text{TTS}$. It is initialized with Qwen2.5-0.5B, and its vocabulary is extended as $\mathbb{V}' = \mathbb{V} \cup \{ <i> \mid i \in \mathbb{N}, 0 \leq i < 6561 \}$, where $\mathbb{V}$ is the original vocabulary. This extension enables the model to generate speech tokens.

The input to $\mathcal{M}_\text{TTS}$ comes from the output of the LLM. Specifically, the LLM output consists of two parts: \textit{continuous hidden states} and \textit{text tokens sampled from the hidden states}. The former contains contextual information, while the latter provides precise textual content. We aim to use both as inputs to the text-to-speech language model. This allows the model to both consider the current context and ensure better alignment with the text response when generating speech tokens. During training, the LLM is trained with teacher forcing, so its output hidden states are denoted as $\mathbf{H} = [\mathbf{h}_1, ..., \mathbf{h}_N]$, where $\mathbf{h}_i = \mathcal{M}_\text{LLM}(X, Y^T_{<i})$. The corresponding text is the ground truth $Y^T = [y^T_1, ..., y^T_N]$. We first use a 2-layer feed-forward network (FFN) to map the hidden states to the embedding dimension of $\mathcal{M}_\text{TTS}$, while also obtaining the text embeddings:
\begin{align}
    \mathbf{e}^{\text{hidden}}_i &= \text{FFN}(\mathbf{h}_i), \\
    \mathbf{e}^{\text{emb}}_i &= \text{Emb}(y^T_i),
\end{align}
where $\text{Emb}(\cdot)$ is the embedding layer of $\mathcal{M}_\text{TTS}$. Afterward, we use an element-wise gate fusion mechanism to combine both representations. Specifically, we compute the gate $\mathbf{g}_i$ as follows:
\begin{equation}
    \mathbf{g}_i = \sigma\left( \mathbf{W}_g \left[ \mathbf{e}_i^{\text{hidden}} \parallel \mathbf{e}_i^{\text{emb}} \right] + \mathbf{b}_g \right),
\end{equation}
where $\parallel$ denotes concatenation, $\sigma$ is the sigmoid function, and $\mathbf{W}_g\in\mathbb{R}^{2d\times d}$ and $\mathbf{b}_g\in\mathbb{R}^d$ are the weight and bias parameters of the gate, and $d$ is the embedding size of $\mathcal{M}_\text{TTS}$. Finally, the fused representation is computed as:
\begin{equation}
    \mathbf{c}_i = \mathbf{g}_i \odot \mathbf{e}_i^{\text{hidden}} + (1 - \mathbf{g}_i) \odot \mathbf{e}_i^{\text{emb}},
\end{equation}
where $\odot$ denotes element-wise multiplication. This fused representations $\mathbf{C} = [\mathbf{c}_1, ..., \mathbf{c}_N]$ are then passed to $\mathcal{M}_\text{TTS}$ for generating speech tokens.

To achieve streaming generation, i.e., to generate speech tokens simultaneously during the LLM's output process, we adopt a ``Read-$\mathcal{R}$-Write-$\mathcal{W}$'' strategy, similar to CosyVoice 2. Specifically, we mix the fused representation $\mathbf{C}$ and the speech tokens $Y^U$ at a predefined ratio $\mathcal{R}:\mathcal{W}$. For every $\mathcal{R}$ fused representations read in, the model generates $\mathcal{W}$ speech tokens. Once all fused representations are read, the model continues to generate the remaining speech tokens until completion. During training, cross-entropy loss is computed only for the generated speech tokens as follows:
\begin{equation}
\mathcal{L}_\text{TTS} = -\sum_{i=1}^{M} \log P(y^U_i | \mathbf{C}_{\leq \min\left(\left\lfloor \frac{i-1}{\mathcal{W}}+1 \right\rfloor \cdot \mathcal{R}, N\right)}, Y^U_{<i}),
\end{equation}
where $\mathbf{C}_{\leq \min\left(\left\lfloor \frac{i-1}{\mathcal{W}}+1 \right\rfloor \cdot \mathcal{R}, N\right)}$ denotes the fused representations that have already been read.

\paragraph{Flow Matching Model}
The speech tokens generated by $\mathcal{M}_\text{TTS}$ are further processed by a chunk-aware causal flow matching model~\citep{lipman2023flow} to synthesize the mel spectrogram in a streaming manner. Every time $\mathcal{W}$ speech tokens are generated, they are treated as a chunk for mel spectrogram synthesis. The synthesized mel spectrogram is then passed through a HiFi-GAN vocoder~\citep{hifigan} to generate the final waveform. We use the pretrained flow matching model and vocoder in CosyVoice 2.

\subsection{Training}
The training of \name relies solely on 200K multi-turn speech-to-speech dialogue data (we will describe how this is synthesized in Section~\ref{sec:data}) and does not use any ASR or TTS data. We find that it is sufficient to achieve excellent performance while minimizing training costs. Specifically, the training process consists of two stages, as shown in Figure~\ref{fig:model}.

\paragraph{Stage I} In Stage I training, we train the speech-to-text and text-to-speech components separately. The training data consists of <speech instruction, text response> pairs and <text response, speech response> pairs from the multi-turn speech-to-speech dialogue data. Specifically, for the speech-to-text part (Stage I(a)), we freeze the speech encoder and train the speech adapter and LLM with cross-entropy loss. For the text-to-speech part (Stage I(b)), we train the text-to-speech language model with cross-entropy loss. Note that during this stage, the gate fusion module is not trained, and only text embeddings are input into $\mathcal{M}_\text{TTS}$.

\paragraph{Stage II}
In Stage II, we train the model's speech-to-speech generation capability with speech-to-speech dialogue data. During this stage, we freeze the speech encoder, speech adapter, and LLM, and only train the gate fusion module and $\mathcal{M}_\text{TTS}$. 

\subsection{Inference}
During inference, the LLM autoregressively generates the text response based on the speech instruction. After generating $\mathcal{R}$ text tokens, its hidden states and the corresponding decoded text are fed into the gate fusion module and $\mathcal{M}_\text{TTS}$ to generate $\mathcal{W}$ speech tokens, which are then passed through the flow matching model and the vocoder to synthesize a speech chunk. In this way, text and speech responses can be generated simultaneously. The response latency for the first synthesized speech chunk can be calculated as:
\begin{equation}
    \mathcal{T}_\text{total} = \mathcal{T}_\text{LLM}(\mathcal{R}) + \mathcal{T}_\text{TTS}(\mathcal{W}) + \mathcal{T}_\text{FM}(\mathcal{W}) + \mathcal{T}_\text{Voc}(2\mathcal{W}),
\label{eq:latency}
\end{equation}
where $\mathcal{T}_\text{LLM}(\mathcal{R})$ and $\mathcal{T}_\text{TTS}(\mathcal{W})$ represent the time required by the $\mathcal{M}_\text{LLM}$ and $\mathcal{M}_\text{TTS}$ models to generate $\mathcal{R}$ and $\mathcal{W}$ tokens, respectively. $\mathcal{T}_\text{FM}(\mathcal{W})$ and $\mathcal{T}_\text{Voc}(2\mathcal{W})$ represent the decoding times of the flow matching model and vocoder when the inputs are $\mathcal{W}$ and $2\mathcal{W}$ tokens\footnote{The length of the mel spectrogram is twice that of the speech tokens (50 Hz vs. 25 Hz).}, respectively.

\section{Data Construction}
\label{sec:data}

In this section, we introduce the process of constructing multi-turn speech-to-speech dialogue data. Our data is an extension of the InstructS2S-200K dataset introduced in~\citet{fang2025llamaomni}, which contains 200K single-turn instruction-following samples designed for speech interaction scenarios. These samples are derived from the Alpaca~\citep{alpaca} and UltraChat~\citep{ultrachat} datasets through rewriting using LLMs. Specifically, for each sample, we first sample the number of turns from a Poisson distribution: $N\sim \text{Poisson}(\lambda=2)$, then clip $N$ to the range of 1 to 5. Next, we use the Llama-3.3-70B-Instruct\footnote{\url{https://huggingface.co/meta-llama/Llama-3.3-70B-Instruct}}~\citep{dubey2024llama3} model to iteratively generate the dialog. For the $i$-th turn, the instruction and response are generated based on the dialogue history of previous $i-1$ turns. In this way, we obtain 200K multi-turn text dialog samples.

Next, we need to convert the text dialogue into speech. To simulate real-world applications, we aim to have varied voices for the instruction, while maintaining a consistent voice for the response. For each multi-turn dialogue, we first use the fish-speech-1.5\footnote{\url{https://huggingface.co/fishaudio/fish-speech-1.5}} model~\citep{fish-speech-v1.4} to synthesize a short prompt (e.g., "This is a randomly generated voice") with a random voice. Then, we use the synthesized speech as the prompt for the CosyVoice2-0.5B\footnote{\url{https://www.modelscope.cn/studios/iic/CosyVoice2-0.5B}} model, which synthesize the instruction into speech while simultaneously cloning the voice. This ensures consistency in the voice across different turns of the dialogue, while maintaining diversity across dialogues. For all responses, we use a uniform voice as the prompt and then synthesize the speech using the CosyVoice2-0.5B model.

\section{Experiments}
\begin{table*}[t]
\centering
\resizebox{\textwidth}{!}{
\begin{tabular}{l|cccc|ccccc}
\toprule
\multirow{3}{*}{\textbf{Model}} & \multicolumn{4}{c|}{\textbf{SpokenQA (Accuracy $\uparrow$)}} & \multicolumn{5}{c}{\textbf{Speech Instruction Following}} \\
\cmidrule{2-5} \cmidrule{6-10}
 & \multicolumn{2}{c}{\textbf{Llama Questions}} & \multicolumn{2}{c|}{\textbf{Web Questions}} & \multicolumn{2}{c}{\textbf{ChatGPT Score $\uparrow$}} & \multirow{2}{*}{\textbf{ASR-WER $\downarrow$}} & \multirow{2}{*}{\textbf{UTMOS $\uparrow$}} & \multirow{2}{*}{\textbf{Latency (ms) $\downarrow$}} \\
 & \textbf{S2T} & \textbf{S2S} & \textbf{S2T} & \textbf{S2S} & \textbf{S2T} & \textbf{S2S} &  &  & \\
\midrule
TWIST & - & 4.0 & - & 1.5 & - & - & - & - & - \\
SpeechGPT & 21.6 & - & 6.5 & - & 2.98 & 2.17 & 40.01 & 3.51 & 5587.94 \\
Spectron & 21.9 & - & 6.1 & - & - & - & - & - & - \\
Moshi (7B) & 62.3 & 21.0 & 26.6 & 9.2 & - & - & - & - & - \\
GLM-4-Voice (9B) & 64.7 & 50.7 & 32.2 & 15.9 & 4.16 & 4.09  & 9.02 & 3.48 & 1562.81 \\
LLaMA-Omni (8B) & 67.7 & 49.0 & 33.4 & 23.7 & 3.99 & 3.52 & 5.95 & 3.67 & \textbf{346.73} \\
\midrule
LLaMA-Omni2-0.5B & 45.7 & 38.7 & 17.7 & 16.8 & 3.24 & 3.20 & \textbf{2.64} & 4.21 & 542.71 \\
LLaMA-Omni2-1.5B & 62.0 & 52.7 & 28.2 & 26.6 & 4.01 & 3.91 & 3.06 & \textbf{4.22} & 552.76 \\
LLaMA-Omni2-3B & 64.3 & 55.7 & 30.5 & 28.0 & 4.24 & 4.14 & 3.37 & \textbf{4.22}  & 567.84 \\
LLaMA-Omni2-7B & 70.3 & 60.7 & 34.5 & 31.3 & 4.28 & 4.15 & 3.26 & 4.19 & 582.91 \\
LLaMA-Omni2-14B & \textbf{73.0} & \textbf{62.7} & \textbf{40.4} & \textbf{37.1} & \textbf{4.56} & \textbf{4.35} & 3.89 & 4.20 & 663.32 \\
\bottomrule
\end{tabular}}
\caption{Results on speech question answering and speech instruction following benchmarks. S2T and S2S represent speech-to-text and speech-to-speech, respectively. We set $\mathcal{R}=3$ and $\mathcal{W}=10$ for all LLaMA-Omni2 series models.}
\label{tab:main}
\end{table*}

\subsection{Experimental Setups}
\paragraph{Model Configuration}
We use the encoder of Whisper-large-v3 as the speech encoder. The speech adapter first performs a 5$\times$ downsampling, followed by a FFN with an intermediate dimension of 2048. For the LLM, we select the Qwen2.5 series models, including Qwen2.5-0.5B/1.5B/3B/7B/14B-Instruct models. We refer to the corresponding models as LLaMA-Omni2-0.5B/1.5B/3B/7B/14B in the following sections. For the text-to-speech language model, we initialize it with the Qwen2.5-0.5B model and set the read-write strategy with $\mathcal{R} = 3$ and $\mathcal{W} = 10$. We will discuss the impact of these hyperparameters on speech quality and response latency later. The speech tokenizer, flow matching model, and vocoder are directly taken from CosyVoice 2.

\paragraph{Training Details}
We use the 200K multi-turn speech-to-speech dialogue data from Section~\ref{sec:data} for two-stage training. In Stage I(a), we freeze the speech encoder and train all parameters of the speech adaptor and LLM. The batch size is 32, and we train for 3 epochs with a peak learning rate of 5e-5. In Stage I(b), we train the text-to-speech language model with a batch size of 32 for 5 epochs and a peak learning rate of 5e-4. In Stage II, we freeze the speech encoder, speech adaptor, and LLM, and train the remaining components with a batch size of 32 for 1 epoch and a peak learning rate of 1e-3. For all stages, we use a warmup strategy for the first 3\% of steps and a cosine annealing learning rate scheduler. The LLaMA-Omni2-14B model is trained on 4 NVIDIA H800 GPUs, while other models are trained on 4 NVIDIA L40 GPUs.

\subsection{Evaluation}
Our evaluation includes two tasks: spoken question answering and speech instruction following. For both tasks, we evaluate the model's speech-to-text and speech-to-speech capabilities. The speech-to-speech evaluation is done by transcribing the speech response into text using the Whisper-large-v3 model, and then applying the same evaluation method as used for speech-to-text evaluation. In all experiments, we use greedy search for the LLM to ensure stable results. For the text-to-speech language model, we use sampling with temperature set to 1.0, as we find that using greedy search causes the model to fall into repetition.

\paragraph{Spoken Question Answering} 
The speech question answering (SpokenQA) task involves asking the model spoken questions, then checking whether the reference answer appears in the model's response, and calculating the accuracy. We evaluate our model on two benchmarks: Llama Questions\footnote{\url{https://github.com/google-research-datasets/LLAMA1-Test-Set}}~\citep{nachmani2024spoken} and Web Questions\footnote{\url{https://huggingface.co/datasets/Stanford/web_questions}}~\citep{berant-etal-2013-semantic}. Since the questions in the Web Questions dataset are in text form, we use CosyVoice2-0.5B to synthesize them into speech.

\paragraph{Speech Instruction Following}
For the speech instruction following task, we follow the settings in~\citet{fang2025llamaomni}, selecting the \textit{helpful\_base} and \textit{vicuna} subsets from the Alpaca-Eval\footnote{\url{https://github.com/tatsu-lab/alpaca_eval}}~\citep{alpacaeval} dataset, excluding math and code-related instructions. The remaining 199 instructions are then synthesized into speech for evaluation. Following~\citet{fang2025llamaomni}, we evaluate the model using the following metrics:

\textbf{ChatGPT Score}: To evaluate the model's ability to follow instructions, we use GPT-4o~\citep{gpt4o} to score the model's responses. It considers factors such as helpfulness, relevance, fluency, and suitability for speech interaction scenarios, and assigns a single score between 1 and 5. The detailed prompt can be found in Appendix~\ref{app:prompt}.

\textbf{ASR-WER}: To assess the consistency between model's text and speech responses, we use Whisper-large-v3 to transcribe the speech response into text, and calculate the word error rate (WER) between the transcribed text and text response. We perform text normalization\footnote{\url{https://github.com/openai/whisper/blob/main/whisper/normalizers/english.py}} before calculating the WER.

\textbf{UTMOS}: To evaluate the naturalness of the generated speech, we use the UTMOS model\footnote{\url{https://github.com/tarepan/SpeechMOS}}~\citep{saeki22c_interspeech} to predict the mean opinion score (MOS) of the generated speech.

\textbf{Latency}: We measure the time from receiving the speech instruction to generating the first speech chunk on a single NVIDIA L40 GPU.

\subsection{Baseline Systems}
We primarily compare \name with the following baseline systems: 

\paragraph{LLaMA-Omni}~\citep{fang2025llamaomni}: One of the earliest SpeechLMs that achieves real-time speech interaction, by using a CTC-based~\citep{ctc} streaming speech decoder to simultaneously generate text and speech units. The generated units are fed into the vocoder for streaming synthesis in fixed-size chunks. We set the chunk size $\Omega = 40$.

\paragraph{GLM-4-Voice}~\citep{glm4voice}: The current state-of-the-art native SpeechLM, pretrained on millions of hours of speech data. It enables real-time speech interaction by alternately generating text and speech tokens in a fixed ratio of 13:26. The generated speech tokens are input into a flow matching model with a fixed chunk size.

In addition, we also borrow some results from~\citet{glm4voice}, including results of TWIST~\citep{hassid2024textually}, SpeechGPT~\citep{zhang-etal-2023-speechgpt}, Spectron~\citep{nachmani2024spoken}, and Moshi~\citep{kyutai2024moshi}.

\section{Results and Analysis}
\subsection{Main Results}

Table~\ref{tab:main} presents the main results on the speech question answering and speech instruction following benchmarks.

\paragraph{Spoken Question Answering}
For the SpokenQA task, we observe that: (1) For models with similar parameter sizes, LLaMA-Omni2-7B outperforms both GLM-4-Voice and LLaMA-Omni in both S2T and S2S settings. Notably, our model significantly reduces the gap between S2T and S2S performance. For example, on the Web Questions benchmark, GLM-4-Voice drops by 16.3 (32.2$\rightarrow$15.9), LLaMA-Omni drops by 9.7 (33.4$\rightarrow$23.7), while LLaMA-Omni2-7B only drops by 3.2 (34.5$\rightarrow$31.3), demonstrating that our approach largely improves speech generation capabilities. (2) For models with varying parameter sizes, we observe that accuracy increases as the LLM size grows, indicating that \name effectively leverages the LLM’s inherent capabilities. For smaller models, LLaMA-Omni2-1.5B/3B exceeds the accuracy of GLM-4-Voice and LLaMA-Omni in the S2S setting, making them suitable choices for edge devices. For larger models, we observe a significant accuracy improvement with LLaMA-Omni2-14B compared to LLaMA-Omni2-7B, highlighting the potential of our approach for scaling to larger models.

\paragraph{Speech Instruction Following}
For the speech instruction following task, we observe that: (1) LLaMA-Omni2-3B/7B/14B outperforms both GLM-4-Voice and LLaMA-Omni in the S2T and S2S settings, demonstrating the strong instruction-following capabilities of our models. (2) Similar to the results on SpokenQA benchmarks, we observe that model performance improves as the LLM size increases, with LLaMA-Omni2-14B achieving significantly better performance. (3) The models’ ASR-WER is generally low, significantly lower than previous models, proving that our models maintain strong consistency between the text and speech responses. (4) Regarding speech quality, thanks to the CosyVoice 2's strong causal flow matching model, our models achieve good UTMOS scores under streaming synthesis, significantly outperforming the baseline models. (5) The latency of \name is around 600ms. Although it is slightly higher than LLaMA-Omni, it still meets the requirements for real-time interaction and is significantly lower than that of GLM-4-Voice.

\begin{table}[t]
    \centering
    \resizebox{\linewidth}{!}{
    \begin{tabular}{l|cc}
        \toprule
       \textbf{Model} & \textbf{Score (S2S)} & \textbf{ASR-WER} \\
       \midrule
        LLaMA-Omni2-7B & \textbf{4.15} & \textbf{3.26} \\
        \quad w/o Gate Fusion & 4.02 & 4.89 \\
        \quad\quad w/o Text Embedding & 3.88 & 6.83 \\
        \bottomrule
    \end{tabular}}
    \caption{Ablation study on the gate fusion module with LLaMA-Omni2-7B.}
    \label{tab:ablation}
\end{table}
\begin{table}[t]
    \centering
    \begin{tabular}{l|cc}
        \toprule
       \textbf{Model} & \textbf{Score (S2S)} & \textbf{ASR-WER} \\
       \midrule
        Streaming TTS & \textbf{4.15} & \textbf{3.26} \\
        Offline TTS & 4.13 & 3.51 \\
        Text Pretrained & 3.53 & 10.34 \\
        Scratch & 1.08 & 80.65 \\
        \bottomrule
    \end{tabular}
    \caption{Ablation study on different TTS pretraining strategies with LLaMA-Omni2-7B.}
    \label{tab:pretrain}
\end{table}
\begin{table}[t]
    \centering
    \resizebox{\linewidth}{!}{
    \begin{tabular}{cc|cccc}
        \toprule
       \textbf{$\mathcal{R}$} & \textbf{$\mathcal{W}$} & \textbf{Score (S2S)} & \textbf{ASR-WER} & \textbf{UTMOS} & \textbf{Latency (ms)} \\
       \midrule
       1 & 5 & 4.09 & 3.48 & 3.98 & \textbf{457.29} \\
       2 & 10 & \textbf{4.15} & 4.00 & 4.19 & 557.79 \\
       3 & 10 & \textbf{4.15} & \textbf{3.26} & 4.19 & 582.91 \\
       3 & 15  & 4.12 & 4.37 & 4.27 & 663.32 \\
       4 & 15 & 4.10 & 3.77 & 4.27 & 683.42 \\
       5 & 20 & \textbf{4.15} & 3.62 & 4.32 & 798.99 \\
       \multicolumn{2}{c|}{Offline} & 4.14 & 3.40 & \textbf{4.46} & - \\
        \bottomrule
    \end{tabular}}
    \caption{Ablation study on the read/write strategy with LLaMA-Omni2-7B. ``Offline'' means generating speech tokens only after receiving the complete input, and then synthesizing all speech tokens into waveform at once.}
    \label{tab:readwrite}
\end{table}

\begin{table*}[t]
\centering
\begin{tabular}{cc|cccc|cccc}
\toprule
\multirow{3}{*}{\textbf{\#Samples}} & \multirow{3}{*}{\textbf{Multiturn}} & \multicolumn{4}{c|}{\textbf{SpokenQA (Accuracy)}} & \multicolumn{3}{c}{\textbf{Speech Instruction Following}} \\
\cmidrule{3-6} \cmidrule{7-9}
 & & \multicolumn{2}{c}{\textbf{Llama Questions}} & \multicolumn{2}{c|}{\textbf{Web Questions}} & \multicolumn{2}{c}{\textbf{ChatGPT Score}} & \multirow{2}{*}{\textbf{ASR-WER}}\\
 & & \textbf{S2T} & \textbf{S2S} & \textbf{S2T} & \textbf{S2S} & \textbf{S2T} & \textbf{S2S} &  \\
\midrule
200K & \checkmark & 70.3 & \textbf{60.7} & 34.5 & 31.3 & \textbf{4.28} & \textbf{4.15} & \textbf{3.26} \\
200K & \texttimes & 70.0 & 59.0 & 33.7 & 30.5 & 4.11 & 3.98 & 3.28 \\
150K & \checkmark & \textbf{70.7} & 58.7 & \textbf{34.7} & \textbf{31.7} & 4.23 & 4.10 & 3.71 \\
100K & \checkmark & 67.7 & 55.3 & 34.1 & 29.9 & 4.19 & 4.07 & 4.45 \\
50K & \checkmark & 50.0 & 37.0 & 16.6 & 13.9 & 3.02 & 2.84 & 5.42 \\
\bottomrule
\end{tabular}
\caption{Results under different training data sizes with LLaMA-Omni2-7B.}
\label{tab:data}
\end{table*}

\subsection{Ablation Studies}
To understand the impact of different factors on overall performance, we conduct a series of ablation studies on the LLaMA-Omni2-7B model.

\paragraph{Gate Fusion Module}
Table~\ref{tab:ablation} shows the ablation study on the gate fusion module. Gate fusion module allows the model to adaptively fuse LLM hidden states and text embeddings, considering both contextual information and textual content. When the gate fusion module is removed and the two components are simply added together ($\mathbf{e}_i^\text{hidden} + \mathbf{e}_i^\text{emb}$) as input to the text-to-speech language model, we observe a decrease in performance. Further removing the text embedding and only inputting the hidden states ($\mathbf{e}_i^\text{hidden}$) results in a further performance decline. This validates the effectiveness of adding text embeddings as input and adaptively fusing them with the gate fusion module.

\paragraph{TTS Pretraining}
Our text-to-speech language model is initialized with the Qwen2.5-0.5B model and undergoes streaming TTS pretraining using text-speech pairs from speech dialogue data in Stage I(b) ($\mathcal{R}=3, \mathcal{W}=10$). We also explore several other strategies, as shown in Table~\ref{tab:pretrain}. ``Offline TTS'' refers to pretraining with the offline TTS task on top of Qwen2.5-0.5B, which shows a slight performance drop compared to the streaming TTS pretraining. ``Text Pretrained'' refers to directly initializing with Qwen2.5-0.5B (with the extended vocabulary including speech tokens), and we observe a significant performance decline. ``Scratch'' refers to a randomly initialized model, whose loss fails to converge within a short period. These experiments demonstrate the importance of pretraining for the TTS language model.

\paragraph{Read/Write Strategy}
The read/write strategies of the TTS language model is a key factor influencing performance, primarily affecting the speech quality and system response latency. As shown in Table~\ref{tab:readwrite}, we explore different combinations of $\mathcal{R}$ and $\mathcal{W}$. First, we observe that when $\mathcal{R} = 3$ and $\mathcal{W} = 10$, the ASR-WER is the lowest, indicating the best alignment between speech and text responses. As for the UTMOS score, we find that it is primarily determined by $\mathcal{W}$, as $\mathcal{W}$ represents the chunk size of speech tokens input to the flow matching model, with larger chunk sizes leading to better speech quality. Regarding response latency, it is jointly determined by $\mathcal{R}$ and $\mathcal{W}$, as shown in Equation~\ref{eq:latency}. Without any engineering optimizations, LLaMA-Omni2-7B can achieve a latency below 500ms. We choose $\mathcal{R} = 3$ and $\mathcal{W} = 10$ in our main experiments because it provides a good trade-off across all aspects.

\subsection{Effects of the Training Data Sizes}

We explore the impact of different training data sizes on performance. As shown in Table~\ref{tab:data}, we first observe that, with the same number of training samples, multi-turn dialogue data consistently achieves better results across all benchmarks compared to single-turn dialogue data, highlighting the effectiveness of multi-turn dialogue data for training. Additionally, for different training data sizes, we observe that as the data size increases, the model's performance improves, gradually stabilizing at 200K training samples. This indicates that our 200K multi-turn dialogue data is generally sufficient while ensuring efficient training.

\section{Related Work}
With the rapid development of LLMs, SpeechLMs have gained widespread attention in recent years~\citep{cui2024recent, ji2024wavchat}, aiming to endow LLMs with the ability to understand or generate speech. Generally speaking, SpeechLMs can be divided into two categories: native SpeechLMs and modular SpeechLMs. Native SpeechLMs refer to decoder-only Transformer models capable of directly inputting and outputting speech tokens. Some early works include SpeechGPT~\citep{zhang-etal-2023-speechgpt, zhang2024speechgptgen}, AudioPaLM~\citep{rubenstein2023audiopalm}, and TWIST~\citep{twist}. These models first convert speech into discrete tokens, then extend the vocabulary of pretrained LLMs to include these tokens, and finally train the LLMs using a large amount of speech or speech-text pair data. Spirit-LM~\citep{nguyen2024spirit} and GLM-4-Voice~\citep{zeng2025scaling, glm4voice} propose training models using speech-text interleaved data to encourage cross-modal knowledge transfer. Moshi~\citep{kyutai2024moshi}, OmniFlatten~\citep{zhang2024omniflatten} and LSLM~\citep{lslm} propose models capable of full-duplex conversations. IntrinsicVoice~\citep{zhang2024intrinsicvoice} proposes a GroupFormer architecture to shorten speech length to be closer to that of text. In contrast to native SpeechLMs, modular SpeechLMs add speech-related modules on top of LLMs. Early works achieve speech understanding tasks by combining speech encoders with LLMs, but are unable to perform speech generation~\citep{wu2023decoder, wang2023blsp, Qwen-Audio, yu2024connecting, slam-asr, hono-etal-2024-integrating,  chen-etal-2024-llast, tang2024salmonn, Qwen2-Audio, fathullah2024audiochatllama}. To achieve speech generation, LLaMA-Omni~\citep{fang2025llamaomni}, Freeze-Omni~\citep{xiong2024freeze}, and OpenOmni~\citep{luo2025openomni} add a speech decoder after LLMs. Mini-Omni~\citep{xie2024miniomni} and SLAM-Omni~\citep{chen2024slamomni} enable LLMs to generate speech tokens simultaneously while generating text tokens. The most related work to ours is the concurrent work Minmo~\citep{chen2025minmo}, which also adopts an autoregressive streaming speech decoder similar to CosyVoice 2. In comparison, Minmo is trained on 1.4M hours of data, while we train on only a few thousand hours of data, providing a more efficient training solution. Additionally, we conduct detailed ablation studies on LLM sizes, read-write strategies, and model architecture to offer a more comprehensive understanding of the model.

\section{Conclusion}
In this paper, we introduce \name, a series of speech language models ranging from 0.5B to 14B parameters, designed to enable real-time, high-quality speech interaction. \name achieves streaming speech generation by integrating an autoregressive text-to-speech language model and a causal flow matching model. Experimental results on spoken question answering and speech instruction following tasks show that \name outperforms previous state-of-the-art speech language models, including LLaMA-Omni and GLM-4-Voice. Additionally, \name can achieve latency under 600ms, meeting real-time interaction requirements. We also conduct detailed ablation studies to understand the impact of various factors on overall performance. In the future, we will explore enhancing \name to generate more human-like speech, incorporating features such as emotion and dialects.

\section*{Limitations}
One limitation of our model is that currently it cannot generate speech responses with different styles (such as emotion or speech rate) based on the content of the input speech or underlying paralinguistic information, as we have only trained on conventional speech-to-speech dialogue data. However, we believe this functionality can be achieved through a data-driven approach, as our model is end-to-end trained and could acquire this capability after further training with suitable data. We plan to explore this in the future.

\section*{Ethical Considerations}
Since \name is built on LLMs, it carries some of the same risks as LLMs, such as the potential for factual errors or other hallucination issues in its outputs. We recommend that the model's outputs be checked in practical use to ensure they comply with the required standards.

\bibliography{custom}
\newpage
\appendix

\section{Prompt}
\label{app:prompt}

\begin{tcolorbox}
[title=Prompt for ChatGPT Scoring (Model: GPT-4o) ,colback=blue!10,colframe=blue!50!black,arc=1mm,boxrule=1pt,left=1mm,right=1mm,top=1mm,bottom=1mm, fonttitle=\small]
\small
I need your help to evaluate the performance of several models in a speech interaction scenario. The models receive the user's speech input and respond with speech output. For evaluation purposes, both the user's speech input and the model's speech response have been transcribed into text using Automatic Speech Recognition (ASR). Your task is to rate the model's responses based on the provided user input transcription [Instruction] and the model's output transcription [Response]. Please consider factors such as helpfulness, relevance, fluency, and suitability for speech interaction in your evaluation, and provide a single score on a scale from 1 to 5.

\vspace{1em}

Below are the transcription of user's instruction and models' response:

\#\#\# [Instruction]: \textbf{\{instruction\}}

\#\#\# [Response]: \textbf{\{response\}}

\vspace{1em}

After evaluating, please output the scores in JSON format: \{score: ...\}. You don't need to provide any explanations.
\end{tcolorbox}

\section{Detailed Latency}

We list the detailed latency at different stages of the model in Table~\ref{tab:latency}. ``LLM'' refers to the latency for generating the first $\mathcal{R}$ text tokens, ``TTS'' refers to the latency for generating the first $\mathcal{W}$ speech tokens, and ``FM+Voc'' refers to the latency for generating the first speech chunk using the flow matching model and vocoder.

\begin{table*}[h]
    \centering
    \begin{tabular}{c|cc|cccc}
        \toprule
         \multirow{2}{*}{\textbf{Model}} & \multirow{2}{*}{$\mathcal{R}$} & \multirow{2}{*}{$\mathcal{W}$} & \multicolumn{4}{c}{\textbf{Latency (ms)}} \\
         & & & \textbf{LLM} & \textbf{TTS} & \textbf{FM+Voc} & \textbf{Total} \\
         \midrule
         LLaMA-Omni2-0.5B & 3 & 10 & 190.95 & 165.83 & 185.93 & 542.71 \\
         LLaMA-Omni2-1.5B & 3 & 10 & 201.01 & 165.83 & 185.93 & 552.76 \\
         LLaMA-Omni2-3B & 3 & 10 & 216.08 & 165.83 & 185.93 & 567.84 \\
         LLaMA-Omni2-7B & 3 & 10 & 231.16 & 165.83 & 185.93 & 582.91 \\
         LLaMA-Omni2-14B & 3 & 10 & 311.56 & 165.83 & 185.93 & 663.32 \\
         \midrule
         LLaMA-Omni2-7B & 1 & 5 & 185.93 & 85.43 & 185.93 & 457.29 \\
         LLaMA-Omni2-7B & 2 & 10 & 206.03 & 165.83 & 185.93 & 557.79 \\
         LLaMA-Omni2-7B & 3 & 10 & 231.16 & 165.83 & 185.93 & 582.91 \\
         LLaMA-Omni2-7B & 3 & 15 & 231.16 & 246.23 & 185.93 & 663.32 \\
         LLaMA-Omni2-7B & 4 & 15 & 251.26 & 246.23 & 185.93 & 683.42 \\
         LLaMA-Omni2-7B & 5 & 20 & 271.36 & 336.68 & 190.95 & 798.99 \\
         \bottomrule
    \end{tabular}
    \caption{Detailed latency of LLaMA-Omni2 series models.}
    \label{tab:latency}
\end{table*}

\end{document}